%% file: root_arxiv.tex
\documentclass[letterpaper, 10 pt, journal, twoside]{IEEEtran}

\title{Improving Pedestrian Prediction Models with Self-Supervised Continual Learning}

\author{Luzia~Knoedler$^*$, Chadi~Salmi$^*$, Hai~Zhu, Bruno~Brito and Javier~Alonso-Mora
\thanks{This paper has received funding from the European Union’s Horizon 2020 research and innovation programme under grant agreement No. 101017008 and Ahold Delhaize. All
content represents the opinion of the author(s), which is not necessarily shared or endorsed by their respective employers and/or sponsors.}
\thanks{The authors are with the Department of Cognitive Robotics, Delft University of Technology, Delft 2628 CD, The Netherlands. \{\tt\footnotesize{l.knoedler; c.salmi; h.zhu; bruno.debrito; j.alonsomora\}@tudelft.nl.}}
\thanks{$^*$Both authors contributed equally to this work.}
\thanks{\blue{\footnotesize{\textbf{Code: \urlstyle{same}\url{https://github.com/tud-amr/scl}}}}}}%

\usepackage{amsmath,amssymb,amsfonts}
\usepackage{algorithmic}
\usepackage{graphicx}
\usepackage{ stmaryrd}
\usepackage{textcomp}
\usepackage{xcolor}
\usepackage{mathtools}
\usepackage{ mathrsfs }
\usepackage{algorithmic}
\usepackage[ruled, linesnumbered]{algorithm2e}
\usepackage{hyperref}
\usepackage{float}
\usepackage{siunitx}
\usepackage{balance}
\usepackage{adjustbox}
\usepackage{makecell}
\usepackage{array}
\usepackage{multirow}
\usepackage{booktabs}
\usepackage{verbatim} 
\usepackage{tabularx}

\usepackage[capitalize]{cleveref}
\usepackage{subcaption}

\makeatletter
\newcommand*{\blue}{\textcolor{black}}

\newcommand*{\rebu}{\textcolor{black}}

\makeatother

\begin{document}

\bstctlcite{IEEEexample:BSTcontrol}

\maketitle


\input{sections/0_abstract}


\input{sections/1_introduction}

\input{sections/2_related_work}

\input{sections/3_problem_formulation}

\input{sections/4_method}
\input{sections/5_results}
\input{sections/6_conclusions}




\bibliographystyle{IEEEtran}
\balance
\bibliography{references}

\end{document}

%% file: sections/0_abstract.tex
\begin{abstract}
Autonomous mobile robots require accurate human motion predictions to safely and efficiently navigate among pedestrians, whose behavior may adapt to environmental changes. 
This paper introduces a self-supervised continual learning framework to improve data-driven pedestrian prediction models online across various scenarios continuously. 
In particular, we exploit online streams of pedestrian data, commonly available from the robot's detection and tracking pipeline, to refine the prediction model and its performance in unseen scenarios.
To avoid the forgetting of previously learned concepts, a problem known as catastrophic forgetting,
our framework includes a regularization loss to penalize changes of model parameters that are important for previous scenarios and retrains on a set of previous examples to retain past knowledge.
Experimental results on real and simulation data show that our approach can improve prediction performance in unseen scenarios while retaining knowledge from seen scenarios when compared to naively training the prediction model online. 
\end{abstract}

%% file: sections/1_introduction.tex
\section{INTRODUCTION}

\IEEEPARstart{A}{utonomous} mobile robots increasingly populate human environments, such as hospitals, airports and restaurants, to perform transportation, assistance and surveillance tasks~\cite{he2017survey}. In these continuously changing environments robots have to navigate in close proximity with pedestrians.
To efficiently and safely navigate around them, robots must be able to reason about human behavior~\cite{pfeiffer2018data}. 
Predicting pedestrian trajectories is challenging, especially in crowded spaces where humans closely interact with their neighbors. This is the case, since the occurring interactions are complex, often subtle, and follow social conventions~\cite{alahi2016social}. Furthermore, humans are influenced by the robot's presence~\cite{trautman2010unfreezing}, features of the static environment, such as its 
geometry or obstacle affordance, and various internal stimuli, such as urgency, which are difficult to measure~\cite{rudenko2020human, lohner2010modeling}.

\begin{figure}[t]
    \centering
    \adjustbox{trim={.0\width} {0\height} {0.00\width} {0\height},clip}{\includegraphics[width=0.35\textwidth]{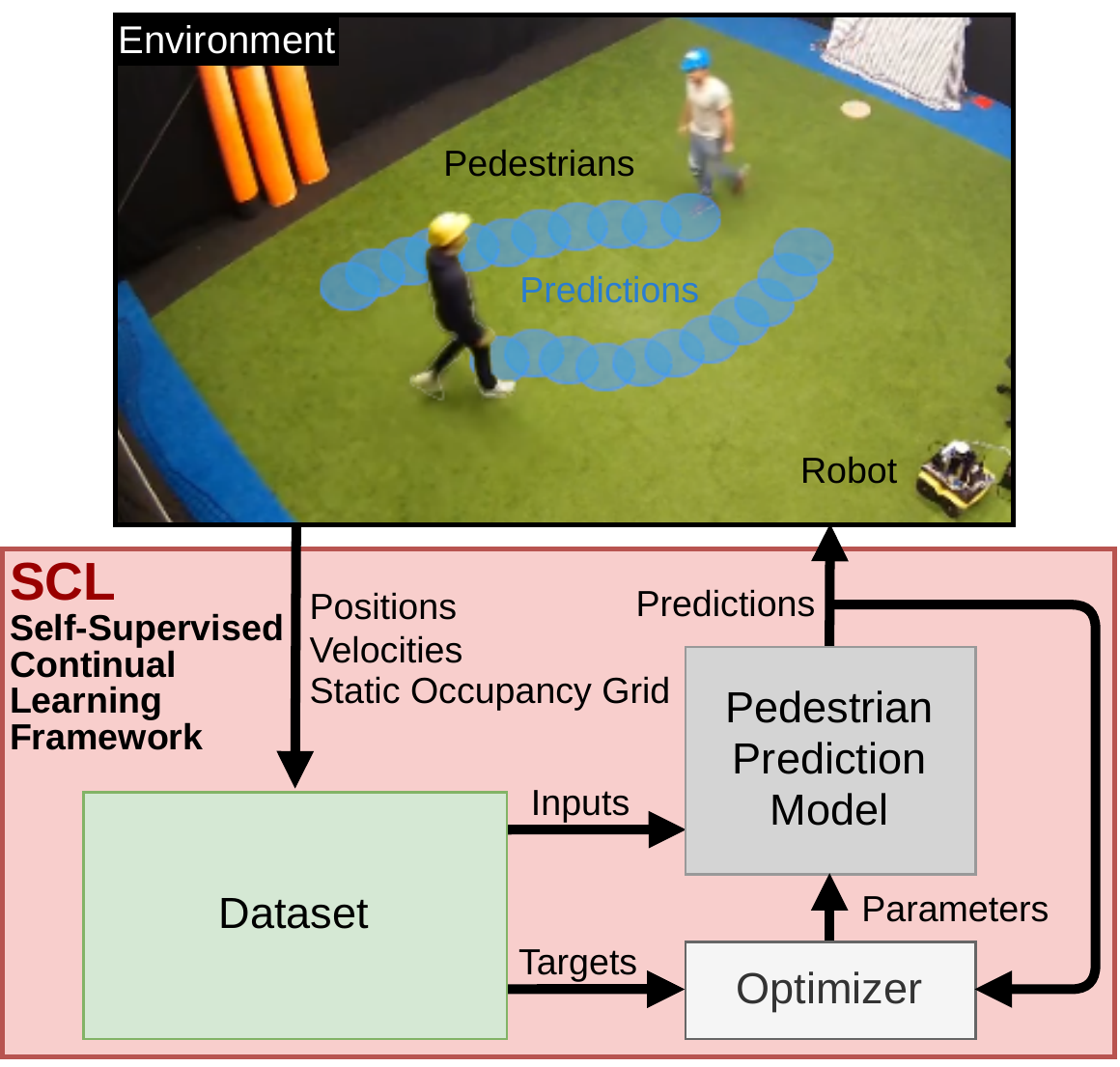}}
    \caption{\blue{Self-supervised Continual Learning~(SCL) framework used to continuously improve data-driven pedestrian prediction models online across various scenarios.}}
    \label{fig:initial_image}
\end{figure}

A large amount of research has been done on pedestrian prediction models~\cite{rudenko2020human}. 
Recently, the focus has mainly been on data-driven models which do not rely on hand-crafted functions and thus allow to capture more complex features and leverage large amounts of data. They address various aspects of pedestrian behaviour such as stochasticity~\cite{ivanovic2019trajectron} and multi-modality~\cite{amirian2019social, brito2019model}. Moreover, they consider the influence of static obstacles~\cite{pfeiffer2018data}, interactions among pedestrians~\cite{alahi2016social} and the robot's presence~\cite{pfeiffer2016predicting}. However, these models are trained offline using supervised learning and thus do not adapt to unseen behaviors or environments and may fail if the testing data distribution differs from the training data distribution.

These limitations can be overcome by continuously training pedestrian prediction models on new streams of data. Hurdles in applying supervised continuous learning to existing prediction models are the slow and expensive creation of labeled data sets or the lack of supervision~\cite{lesort2020continual}. 
Robots operating in the same environment as pedestrians can autonomously collect training examples based on the robot's never-ending stream of observations. If a robot can efficiently and autonomously collect examples, its internal prediction models can be updated on the fly and the robot can effectively adapt its behavior. 
However, neural networks are prone to forget previously learned concepts while sequentially learning new concepts~\cite{lesort2020continual}. This phenomenon is referred to as \textit{catastrophic forgetting}. To overcome catastrophic forgetting, we use a \textit{regularization} strategy, namely elastic weight consolidation~(EWC)~\cite{kirkpatrick2017overcoming}, to selectively slow down learning for important model parameters, in combination with \textit{rehearsing} a small set of examples from previous tasks.

The main contribution of this work is therefore the introduction of a self-supervised continual learning framework that uses online streams of data of pedestrian trajectories to continuously refine data-driven pedestrian prediction models, see \cref{fig:initial_image}. Our approach overcomes catastrophic forgetting by combining a regularization loss and a data rehearsal strategy.

We evaluate the proposed method in simulation, showing that our framework can improve prediction performance over baseline methods and avoid catastrophic forgetting, and in experiments with a mobile robot, showing that our framework can continuously improve a prediction model without the need for external supervision.

%% file: sections/2_related_work.tex
\section{RELATED WORK}

In this section, we describe relevant approaches for pedestrian motion prediction and continual learning.

\subsection{Pedestrian Motion Prediction}
There has been a vast amount of work devoted to pedestrian trajectory prediction~\cite{rudenko2020human}. 
Early works are mainly model-based, such as the well-known social force model (SFM) which uses attracting and repulsive potentials to model the social behaviours of pedestrians \cite{helbing1995social}, and the velocity-based models which compute collision-free velocities for trajectory prediction~\cite{kim2015brvo,bera2016glmp}. A limitation of these model-based approaches is that they only utilize handcrafted features, thus not being able to capture complex interactions in crowded scenarios. 
To overcome the limitation, recurrent neural networks (RNNs) have been used for human trajectory prediction, which allows to represent complex features and leverage large amounts of data~\cite{becker2018evaluation}. Building on RNNs, \cite{alahi2016social} utilized LSTM networks to model time dependencies and employed a pooling layer to model interactions. \cite{pfeiffer2016predicting} proposed a network model that is aware of the environment constraints. In addition, other network models have been developed to predict pedestrian trajectories, including Generative Adversarial Networks (GANs)~\cite{amirian2019social, gupta2018social} and Conditional Variational Autoencoders (CVAEs)~\cite{lee2017desire,brito2020social}.
Albeit being efficient, these models are usually trained and evaluated using (offline) bench-marking datasets \cite{pellegrini2009you, lerner2007crowds, caesar2020nuscenes, chang2019argoverse}, which limits their online adaption to unseen scenarios. In this paper, we propose an approach to improve these models online by introducing a self-supervised continual learning framework.

\subsection{Continual Learning}
Continual learning (CL)
addresses the training of a model from a continuous stream of data containing changing input domains or multiple tasks~\cite{delange2021continual}.
The goal of CL is to adapt the model continually over time while preventing new data from overwriting previously learned knowledge. 
Existing CL approaches that mitigate catastrophic forgetting for neural network-based models can be divided into three categories: architecture-, memory- and regularization-based~\cite{lesort2020continual, parisi2019continual}.

\textbf{Architecture-based} approaches change the architecture of the neural network by introducing new neurons or layers~\cite{yoon2017lifelong, hung2019compacting, li2019learn}. Intuitively, these approaches prevent forgetting by populating new untouched weights instead of overwriting existing ones. However, the model complexity grows with the number of tasks.

\textbf{Memory-based} approaches save samples of past tasks to rehearse previous concepts periodically~\cite{lesort2020continual}. There are two types of memory-based methods that differ in the way they memorize past experiences: \textit{rehearsal} methods explicitly saving examples~\cite{rebuffi2017icarl} and \textit{pseudo-rehearsal} methods saving a generative model from which samples can be drawn \cite{shin2017continual}. The data stored in the memory of rehearsal methods can be randomly chosen or carefully selected~\cite{rebuffi2017icarl, aljundi2019gradient}. Some methods require task boundaries~\cite{rebuffi2017icarl} while other methods can be applied to the task free setting~\cite{aljundi2019gradient}. Since memory-based approaches require a separate memory, they can become unsustainable with an increasing number of tasks.

\textbf{Regularization-based} approaches add a regularization term to the loss to prevent modification of model parameters. This can be done using basic regularization techniques, such as weight sparsification, early stopping, and dropout, or with more complex methods which selectively prevent changes in parameters that are important to previous tasks~\cite{lesort2020continual}.
\cite{kirkpatrick2017overcoming} introduced \textit{Elastic Weight Consolidation}~(EWC), a regularization approach limiting the plasticity of specific neurons based on their importance determined from the diagonal of the Fisher Information Matrix (FIM). To compute the FIM, clear task boundaries are required. Other regularization approaches focus on relaxing this assumption by automatically inferring task-boundaries \cite{aljundi2019task}, or by calculating the importance in an online fashion over the entire learning trajectory~\cite{zenke2017continual}.
In contrast to other categories of approaches, these regularization-based methods do not require much computational and memory resources. However, one downside of regularization-based approaches is that an additional loss term is added, which can lead to a trade-off between knowledge consolidation and performance on novel tasks. 

Most of the time, combining different continual learning strategies results in better performance~\cite{lesort2020continual}.
Hence, in this paper, we employ the EWC regularization technique combined with a data rehearsal strategy to achieve continual learning to improve pedestrian prediction models.

%% file: sections/3_problem_formulation.tex
\section{PROBLEM FORMULATION}

Throughout this paper, we denote vectors, $\mathbf{x}$, in bold lowercase letters, matrices, $M$, in uppercase letters, and sets, $\mathcal{X}$, in calligraphic uppercase letters. 

We address the problem of continuously improving a trajectory prediction model online using \blue{streams of pedestrian data}. This data includes the position and velocity of all $n$ tracked pedestrians over time, and an occupancy map of the static environment $\mathcal{S}$.
The position, velocity, and the surrounding static environment of the $i$-th pedestrian at time~$t$ are denoted by $\mathbf{p}_t^i = [p_{\mathrm{x},t}^i,p_{\mathrm{y},t}^i]$, $\mathbf{v}_t^i = [v_{\mathrm{x},t}^i,v_{\mathrm{y},t}^i]$, and $\mathcal{O}_t^{\mathrm{env}, i} \subset \mathcal{S}$, respectively. The sub-scripts $\mathrm{x}$ and $\mathrm{y}$ indicate the x and y direction in the world frame. 
The super-script $i$ denotes the $\textit{query-agent}$, i.e., the pedestrian whose trajectory we want to predict.

Denote by $\mathcal{X}_{t}^i$ the observations acquired within a past time horizon $t_{\mathrm{obs}}$ for predicting pedestrian $i$'s future trajectory, which typically includes its own states, the states of the other pedestrians and environment information. Further denote by $\hat{\mathcal{Y}}_t^i$ the predicted trajectory of pedestrian $i$ over the future prediction horizon $t_{\mathrm{pred}}$. 

We seek a data-driven prediction model \blue{$\hat{\mathcal{Y}}_{t}^i = f_{\mathbf{\theta}}(\mathcal{X}_{t}^i$}), with parameters $\mathbf{\theta}$, that best approximates the true trajectory~$\mathcal{Y}_{t}^i$ across the entire previous stream of states for every tracked pedestrian $i \in \{1, \ldots , n\}$.
The true trajectory $\mathcal{Y}_{t}^i$ will only become available in hindsight after observing the trajectory taken by pedestrian $i$ during $t_\mathrm{pred}$.
Thus, we formulate the problem of continually learning a data-driven prediction model from past observations at time $t$ as a regret minimization problem:
\begin{equation}
\rebu{\min_{\theta} \sum_{i=1}^{n}
\sum_{\tau=t-t_{\mathrm{his}}}^t \mathscr{L_\mathrm{pred}}(\hat{\mathcal{Y}}_{\tau}^i, \mathcal{Y}_{\tau}^i),}
\end{equation}
where $t_{\mathrm{his}}$ is the entire elapsed time until $t$ and $\mathscr{L_\mathrm{pred}}(\mathcal{\hat{Y}}_{\tau}^i, \mathcal{Y}_{\tau}^i)$ is the regret at one past time step $\tau$ for pedestrian~$i$, which will be described in later sections.

%% file: sections/4_method.tex
\section{METHOD}\label{sec:method}

\begin{figure*}[htbp]
    \centering
    \includegraphics[width=0.80\textwidth]{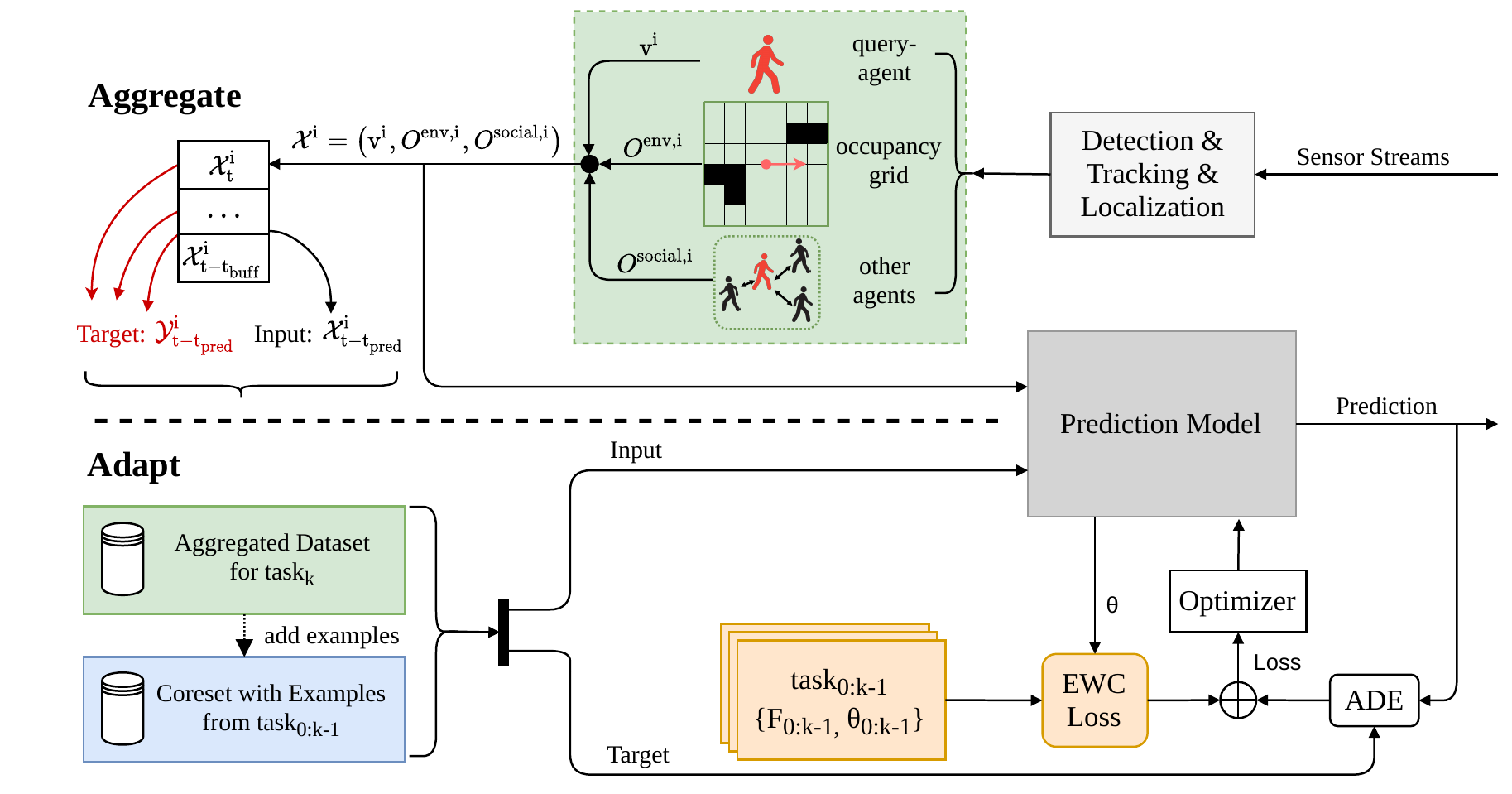}
    \caption{Schematics of the SCL framework. The aggregation dataset is collected by extracting examples from the stream of tracked surrounding pedestrians (task aggregation). The prediction model is trained using the aggregated dataset and a separately saved coreset applying a EWC regularization to prevent catastrophic forgetting (model adaption).}
    \label{fig:framework}
\end{figure*}

In this section, we introduce the Self-supervised Continual Learning (SCL) framework, an online learning framework to continually improve pedestrian prediction models. \cref{SCL} presents the overall structure of SCL, \cref{network_model} the prediction network architecture, \cref{data_aggregation} the data aggregation and \cref{training} the model adaption.

\subsection{Self-supervised Continual Learning}\label{SCL}
\blue{The SCL architecture consisting of two phases: a task aggregation and a model adaptation phase is presented in \cref{fig:framework}}. Firstly, we use a prediction model which was pre-trained on publicly available datasets \cite{pellegrini2009you, lerner2007crowds} and aggregate new training examples using the surrounding pedestrians as experts (task aggregation) for a period of time $T$. Then, we update the prediction model using the aggregated data of the current task and a small constant sized coreset, which contains examples from previous tasks (model adaptation). During the model adaption phase, we apply a EWC loss to preserve the prediction performance on previous tasks. The two phases run alternately over time to create a continuous learning autonomous robot.
During the task aggregation phase, we associate a new task to a new  environment on which the model was previously not trained on.
To distinguish between tasks, we will refer to the currently considered task as $\textrm{task}_k$. The previous tasks are referred to as $\textrm{task}_{0:k-1}$ where the subscript $0$ refers to the initial task.

\begin{figure}[htpb]
    \centering
    \adjustbox{trim={.02\width} {0\height} {0.12\width} {0\height},clip}{\includegraphics[width=0.55\textwidth]{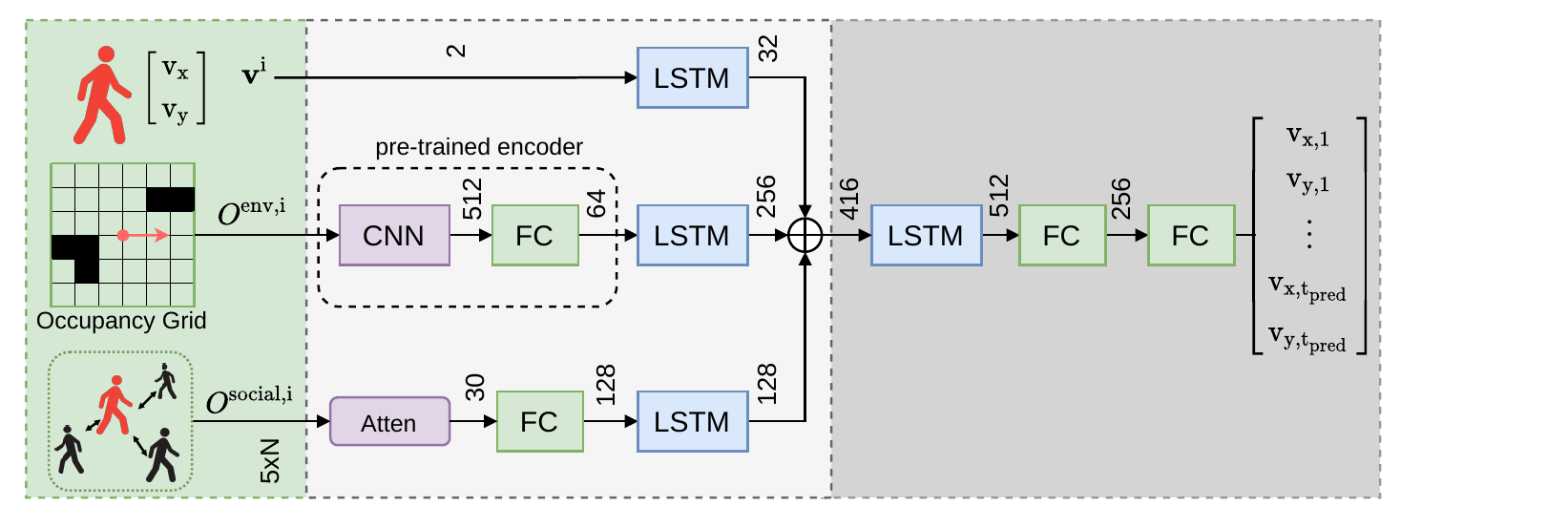}}
    \caption{Pedestrian motion prediction model architecture.}
    \label{fig:model}
\end{figure} 

\subsection{Prediction Network Architecture}\label{network_model}
To evaluate our online learning framework we use a data-driven pedestrian prediction model building on \cite{pfeiffer2018data}. Please note that our approach does not depend on which network model we use. \blue{However, the memory requirements scale linearly with the number of tasks and model parameters.} \blue{\Cref{fig:model}} shows the network model which uses three streams of information. The first input is the query-agent's velocity \blue{over an observation time window $t_\mathrm{obs}$, $\mathbf{v}_{t-t_\mathrm{obs}:t}^i$}, which enables the model to capture the pedestrian's dynamics. 
The second input is the occupancy grid information \blue{$O^\mathrm{env, i}_{t-t_\mathrm{obs}:t}$} that contains information about the static obstacles centered on the query-agent. In contrast to \cite{pfeiffer2018data}, the third input is a vector containing information about the relative position and velocity of surrounding pedestrians \blue{$O^\mathrm{social, i}_{t-t_\mathrm{obs}:t}$}. 
This adaption was done because the model using an angular pedestrian grid, presented in \cite{pfeiffer2018data},  has shown difficulties learning social interactions~\cite{brito2020social}. 
For one neighbor pedestrian $j$ the vector including the relative measurements to the query-agent $i$ at time $t$ is
$$
\mathbf{e}^{i,j}_t = [\mathbf{p}_t^j - \mathbf{p}_t^i, \mathbf{v}_t^j- \mathbf{v}_t^i].
$$
Thus, the information vector at time $t$ is defined as
$$ 
O^\mathrm{social, i}_t = [\mathbf{e}^{i,1}_t, \ldots, \mathbf{e}^{i,i-1}_t, \mathbf{e}^{i,i+1}_t, \ldots, \mathbf{e}^{i,n}_t].
$$
\blue{
Hence, the information used for trajectory prediction of pedestrian $i$ is $\mathcal{X}_t^i = (\mathbf{v}_{t-t_\mathrm{obs}:t}^i, O_{t-t_\mathrm{obs}:t}^\mathrm{env, i}, O_{t-t_\mathrm{obs}:t}^\mathrm{social, i})$, and the prediction model is given by $\hat{\mathbf{v}}_{t+1:t+t_{\mathrm{pred}}}^i = f_{\mathbf{\theta}}(\mathbf{v}_{t-t_\mathrm{obs}:t}^i, O_{t-t_\mathrm{obs}:t}^\mathrm{env, i}, O_{t-t_\mathrm{obs}:t}^\mathrm{social, i})$, where the trajectory predictions are represented by a sequence of velocities, i.e. $\hat{\mathcal{Y}}_t^i = \hat{\mathbf{v}}_{t+1:t+t_{\mathrm{pred}}}^i$.}
We use the permutation invariant $sort$ function as an attention mechanism by sorting the relative vectors of surrounding agents by euclidean distance~\cite{sadeghian2019sophie}. To handle a variable number of pedestrians, only the closest $n$ pedestrians are considered. For situations with fewer than $n$ surrounding pedestrians, the relative vector of the closest pedestrian is repeated. 

\subsection{Task Aggregation}\label{data_aggregation}

For each task $k$, SCL saves the inputs of the prediction model, \rebu{$\mathcal{X}_t^i = (\mathbf{v}_{t-t_\mathrm{obs}:t}^i, O_{t-t_\mathrm{obs}:t}^\mathrm{env, i}, O_{t-t_\mathrm{obs}:t}^\mathrm{social, i})$}, in a buffer for each \blue{time step $t= \{-t_\mathrm{buff}, \ldots, 0\}$} (see \cref{fig:framework}).
Then, for each time step $t$, the ground truth velocity sequence \blue{$\mathbf{v}^i_{t+1:t+t_\mathrm{pred}}$} is extracted in hindsight from the buffer and the corresponding input to the prediction model \blue{$\mathcal{X}_t^i$} (red arrows in \cref{fig:framework}). 
We aggregate the velocity vectors (Target) together with the corresponding model inputs (Input) 
and store them in the aggregated task dataset $\mathcal{D}_k$ \blue{as an example}. The examples are aggregated as a sequence. As we use a recurrent prediction model and train the model with truncated back-propagation through time $t_\mathrm{tbptt}$, we only aggregate sequences \blue{of examples} with a length of \blue{$t_\mathrm{buff} = t_\mathrm{pred} + t_\mathrm{tbptt}$}. We collect training examples for $T$ seconds. 

\subsection{Model Adaption}\label{training}
We present the overall SCL procedure in Algorithm~\ref{algo:framework}. \blue{For each task, we aggregate a dataset $\mathcal{D}_k$ over $T$ seconds. Then, the model is adapted using $\mathcal{D}_k$ and a set containing examples of previous tasks referred to as coreset $\mathcal{D}_\mathrm{coreset}$. The \textit{Coreset Rehearsal} strategy is applied to mitigate forgetting.  
Thus, the training dataset is defined as follows:
$$\hat{\mathcal{D}} = \mathcal{D}_k \bigcup \mathcal{D}_\mathrm{coreset}.$$
In the model adaptation phase, SCL uses the training dataset~$\hat{\mathcal{D}}$ to train the network for $Q$ epochs. }
The training loss is composed by a prediction loss and a regularization loss to avoid catastrophic forgetting: $\mathscr{L_\mathrm{train}} = \mathscr{L}_\mathrm{pred} + \mathscr{L}_\mathrm{reg}$. We define the prediction loss as the average norm between the predicted velocity sequence and the ground truth:
\begin{align}
\begin{split}
 \mathscr{L}_\mathrm{pred}(\hat{\mathcal{Y}}_t^i, \mathcal{Y}_t^i) = \frac{1}{t_\mathrm{pred}} \sum_{\tau=t+1}^{t+t_\mathrm{pred}} | \hat{\mathbf{v}}_{\tau}^i - \mathbf{v}_{\tau}^i |^2.
\end{split}
\end{align}
We employ EWC \cite{kirkpatrick2017overcoming} as regularization loss method to preserve prediction performance on the previous tasks ($\textrm{task}_{0 : k-1}$) and overcome catastrophic forgetting. 
EWC penalizes the distance between the new model parameters, $\mathbf{\theta}$, and the previous task parameters,~$\mathbf{\theta}_{0:k-1}$, depending on their importance to keep the knowledge of previous tasks.
After learning each task, EWC computes the corresponding importance parameter by using the diagonal elements of the FIM $F$, which are defined as:
\begin{equation} \label{eq:fisher}
    F_{k,jj} = \frac{1}{|\mathcal{D}_k|}\sum_{\mathcal{X}\in \mathcal{D}_k} \left(\frac{\delta \log \blue{f_{\mathbf{\theta}}(\mathcal{X})}}{\delta\theta_j}\bigg|_{\mathbf{\theta}=\mathbf{\theta}_k^*} \right)^2,
\end{equation}
where $k$ represents the task number, $\mathcal{D}_k$ is the training data containing trajectories from task $k$, \blue{$f_{\mathbf{\theta}}(\mathcal{X})$ is the predicted output of the network with parameters $\mathbf{\theta}$ given data $\mathcal{X} \in \mathcal{D}_k$}.
The importance measure $F_{k}$ is saved together with the network weights $\mathbf{\theta}_k$.
Based on $F_{0:k-1}$ and $\mathbf{\theta}_{0:k-1}$ the following regularization term is added to the loss function:
\begin{equation}
\mathscr{L}_\mathrm{reg}(\theta) = \sum_{l=0}^{k-1} \frac{\lambda}{2} F_l (\theta - \theta_{l})^2,
\end{equation}
where $\mathbf{\theta}$ is the current set of weights for the current task $k$ and $\lambda$ is the hyperparameter that dictates how important not forgetting the old task is compared to learning the new one.\\
After the model adaptation phase is completed, we update the coreset with $M$ examples of the latest task ($\textrm{task}_k)$. Importantly the new examples replace existing ones to ensure the coreset remains of constant length $N$. We randomly select which examples to drop to update the coreset. After training the datasets $\mathcal{D}_k$ and $\hat{\mathcal{D}}$ are cleared.

\begin{algorithm}[htpb]
\SetAlgoLined
\small
Load pretrained model: \blue{$f_{\mathbf{\theta}}$} \\
Load map: $\mathcal{S}$ \\ 
Initialize coreset: $\mathcal{D}_\mathrm{coreset} \leftarrow \emptyset$ \\
\For{k = 0 to $\infty$}{
    Initialize the empty task dataset: $\mathcal{D}_k \leftarrow \emptyset$ \\
    Aggregate examples for $T$ seconds as follows: \\
    \For{t = 0 to T}{
        Process pedestrian positions $\mathbf{p}_t^{i}$, velocities $\mathbf{v}_t^{i}$, and the occupancy grid $O_t^\mathrm{env, i}$ to model inputs $\mathcal{X}_t^{i}$ and save them to a buffer for $i \in \{1, \ldots , n\}$ \\
        Get examples from buffer: $\mathcal{E}_t = \{(\mathcal{X}_{t}^{1}, \mathcal{Y}_t^1), \ldots, (\mathcal{X}_{t}^{n}, \mathcal{Y}_t^n)\}$ \\
        Update task dataset: $\mathcal{D}_k \leftarrow \mathcal{D}_k \bigcup \mathcal{E}_t$ \\
    }
    Combine coreset and task: $\hat{\mathcal{D}} \leftarrow \mathcal{D}_k \bigcup \mathcal{D}_{coreset}$ \\
    Train prediction model \blue{$f_{\mathbf{\theta}}$} on $\hat{\mathcal{D}}$ using EWC\\
    Save EWC importances $F_k$ and the updated parameters $\mathbf{\theta}_k$ of $\text{task}_k$ \\
    Update coreset $\mathcal{D}_{coreset}$ with $M$ random examples from $\mathcal{D}_k$ \\
    Clear $\mathcal{D}_k$, $\hat{\mathcal{D}}$ from memory \\
}
\caption{The Self-supervised Continual Learning (SCL) framework}
\label{algo:framework}
\end{algorithm}

%% file: sections/5_results.tex
\section{RESULTS}
In this section, we present \blue{quantitative and} qualitative results in both simulation and real-world experiments.

\subsection{Experimental Setup}
The prediction model parameters are displayed in \cref{fig:model}. \blue{We pre-train the prediction model on the ETH and UCY pedestrian datasets~\cite{pellegrini2009you, lerner2007crowds} for 60 epochs. 
Our online learning framework will improve this pre-trained model based on the behavior of surrounding pedestrians.} The applied hyperparameters are summarized in \cref{hp}. Note that although $t_\mathrm{obs}=0$, the past states are implicitly taken into account through the internal memory of the LSTMs.
First, we evaluate our framework in simulation assuming full knowledge of the map and current states of all pedestrians. 
The pedestrian behaviour is simulated using the SFM~\cite{helbing1995social} \rebu{and Reciprocal Velocity Obstacle model~(RVO)~\cite{van2008reciprocal}}. We train the prediction model incrementally on arbitrary orders of these environments.\\
\blue{To evaluate how well our framework scales to complex scenarios with more pedestrians we rerun the above experiments in simulation with an increased number of pedestrians.}\\
Then, we apply SCL in real-world experiments. Here, the true pedestrian behavior differs from the models assumed during simulation.
To eliminate the perception-related errors as much as possible, we first test our framework with an optical tracking system (Optitrack) that provides pose information of all tracked pedestrians. We set up three scenarios to replicate the simulation environments. Finally, we evaluate our framework in an uncontrolled hall using only the on-board sensing and, a detection and tracking pipeline. 

\subsection{Baseline Methods}
We evaluate our method against three baseline approaches in both simulation and real-world experiments:
\begin{enumerate}
    \item \textbf{Offline:} The prediction model is trained offline on all tasks. This baseline represents a performance upper-bound assuming that all data is available.
    \item \textbf{Vanilla:} The prediction model is trained using only the aggregated data and standard gradient descent without any regularization loss.
    \item \textbf{EWC:} The prediction model is trained using only the aggregated data with EWC regularization, but without coreset rehearsal.
\end{enumerate}

\blue{In simulation, we additionally consider the following baselines:
\begin{itemize}
    \item \textbf{Coreset:} The prediction model is trained using the aggregated data and coreset data.
    \item \textbf{CV:} The human behaviour is predicted using the constant velocity model, no learning is applied.
\end{itemize}}
\blue{
The CV model was added since it was shown to outperform state-of-the-art data-based prediction models~\cite{scholler2020constant} and to enable robust navigation around humans~\cite{mavrogiannis2018social}. A limitation of the CV model is that it does not consider obstacles.}

Since our focus is on applying continual learning strategies to improve pedestrian prediction models on the fly without forgetting, we only change the learning strategy across baselines and keep the prediction network architecture fixed. Similar to other works on pedestrian prediction models, we use the average displacement (ADE) and final displacement error (FDE) as performance metrics~\cite{brito2020social, sadeghian2019sophie}.
 
\begin{table}[t]
\renewcommand{\arraystretch}{1}
\caption{Hyperparameters.}\label{hp}
\centering
\begin{tabular}{|p{2.15cm}|p{0.45cm}||p{3.35cm}|p{1.0cm}|}
 \hline
 time-step   & \SI{0.2}{\second} &  \# training epochs $Q$ & \SI{250}{}\\ 
 \hline
 task length $T$ & \SI{200}{\second} & learning rate   & \SI{2E-3}{}\\ 
 \hline
 buffer size $t_\mathrm{buff}$ & \SI{6}{\second} & L2 regularization & \SI{5e-4}{} \\ 
 \hline
 predict. time $t_\mathrm{pred}$ & \SI{3}{\second} & EWC parameter $\lambda$ & \SI{1e6}{}\\ 
 \hline
tbptt time $t_\mathrm{tbptt}$& \SI{3}{\second} & coreset size $N$/update size $M$ & \SI{100}{}/ \SI{20}{}\\ 
 \hline
observ. time $t_{\mathrm{obs}}$  & \SI{0}{\second} & validation set size $L_v$ & \SI{100}{}\\ 
 \hline
\end{tabular}
\end{table}

\subsection{Tasks}
\begin{figure}[htbp]
    \centering
    \includegraphics[width=0.45\textwidth]{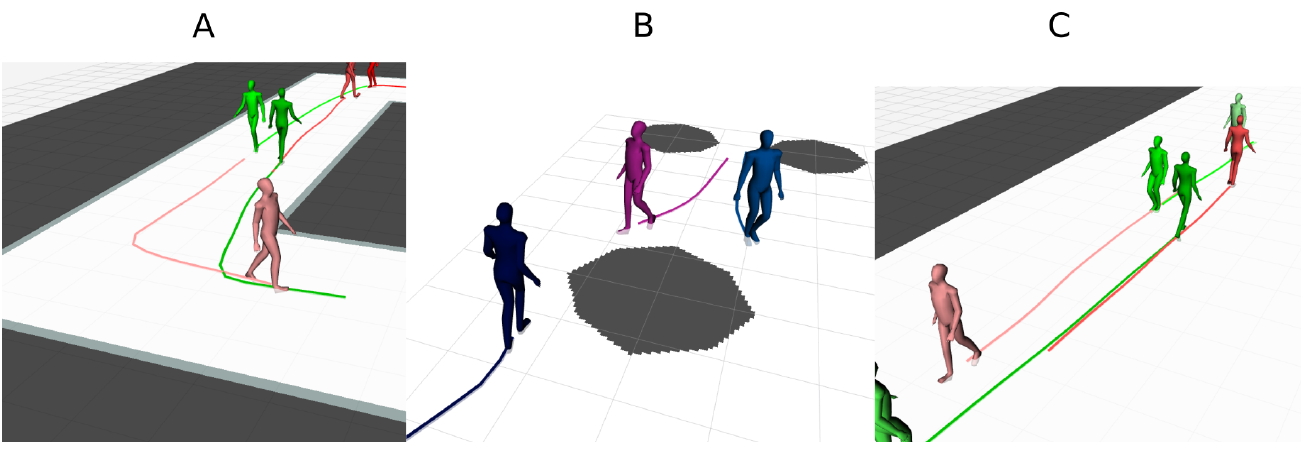}
    \caption{The considered simulation environments consist of (A) Square, (B) Obstacle, and (C) Hall environments.}
    \label{fig:IL-envs}
\end{figure}

We consider three distinct environments, i.e., tasks, displayed in \cref{fig:IL-envs}:
\begin{enumerate}
    \item \textbf{Square:} An infinite corridor setting with three pedestrians walking clockwise and three anticlockwise.
    \item \textbf{Obstacles:} Pedestrians walking towards each other in an obstacle filled space.
    \item \textbf{Hall:} Pedestrians walking towards each other in a hall while behaving cooperatively.
\end{enumerate}
\blue{The scenarios were selected since they include encounters typically experienced in everyday situations. The specific environments were chosen to investigate social interactions (Hall), obstacle interactions (Obstacle) and semantic knowledge of the map (Square).}
\blue{To additionally evaluate our framework in scenarios with more interacting agents we consider the above environments with $10$ and $20$ pedestrians.}
\blue{For the obstacle-free environments, we use the open-source pedsim simulation framework\footnote{\url{https://github.com/srl-freiburg/pedsim_ros}} employing the SFM~\cite{helbing1995social} to simulate the pedestrian behavior. For environments with static obstacles, we employ the RVO method~\cite{van2008reciprocal}\footnote{\url{https://github.com/sybrenstuvel/Python-RVO2}} as pedestrians following the SFM may still collide with obstacles. }

\subsection{Simulation Results}
We evaluate the prediction performance of the network model trained with our method (SCL) versus the baselines on different sequences of environments (square, obstacle, hall) starting from a pre-trained model. Each environment is observed for $T$ seconds to create the aggregated dataset on which the model is trained. Thus, each environment corresponds to a new task. To compute the ADE/FDE performance metrics, we collect a validation set for each environment \blue{including $L_v$ examples} not used during training.

\renewcommand{\arraystretch}{0.8}{
\begin{table*}[t]
\caption{Quantitative results of \blue{the CV prediction and} Vanilla, EWC, \blue{Coreset} and SCL training approaches for four environment sequences. \blue{The results for the dense scenarios
are included as SCL-10 and SCL-20.} \blue{The table lists the mean$\pm$standard deviation (std) of ADE / FDE for all environments at the sequence end under \textit{seq. end}} and the \blue{mean$\pm$std} of the \textit{forgotten} ADE / FDE, which refers to the average increase in ADE / FDE on previous environments across the learning sequence. \blue{All error measures are presented in meters.}}
\label{table:orders-sim}
\centering
\begin{tabular}{|l || l | l || l | l || l | l || l | l |}
\hline
\rule{0pt}{7pt}
 & \multicolumn{2}{c||}{square $\shortrightarrow$ obstacle $\shortrightarrow$ hall} & \multicolumn{2}{c||}{obstacle $\shortrightarrow$ square $\shortrightarrow$ hall} & \multicolumn{2}{c||}{hall $\shortrightarrow$ obstacle $\shortrightarrow$ square} & \multicolumn{2}{c|}{obstacle $\shortrightarrow$ hall $\shortrightarrow$ square}\\
\hline
    Method & \thead{forgotten \\ (mean$\pm$std)}  & \thead{seq. end \\ (mean$\pm$std)} &  \thead{forgotten \\(mean$\pm$std)} & \thead{seq. end \\(mean$\pm$std)} &  \thead{forgotten \\ (mean$\pm$std)}  & \thead{seq. end \\(mean$\pm$std)} &  \thead{forgotten \\(mean$\pm$std)}  & \thead{seq. end \\(mean$\pm$std) }\\
\hline
\rule{0pt}{7pt}
\multirow{2}{*}{\rebu{CV}} & \rebu{+0.00$\pm$0.00/} & \rebu{1.12$\pm$1.18/} & \rebu{+0.00$\pm$0.00/} & \rebu{1.25$\pm$1.29/} & \rebu{+0.00$\pm$0.00/} & \rebu{1.12$\pm$1.17/} &  \rebu{+0.00$\pm$0.00/} & \rebu{1.26$\pm$1.31/}  \\
        & \rebu{+0.00$\pm$0.00} &  \rebu{1.09$\pm$1.69} & \rebu{+0.00$\pm$0.00}& \rebu{1.10$\pm$1.65 }& \rebu{+0.00$\pm$0.00 }& \rebu{1.16$\pm$1.91  }& \rebu{+0.00$\pm$0.00}& \rebu{1.24$\pm$1.92}\\
\rule{0pt}{8pt}        
\multirow{2}{*}{Vanilla}  & +0.12\rebu{$\pm$0.29}/  & 0.21\rebu{$\pm$0.31}/ & +0.10\rebu{$\pm$0.23}/ & 0.21\rebu{$\pm$0.27}/ & +0.20\rebu{$\pm$0.27}/ & 0.27\rebu{$\pm$0.26}/ &  +0.18\rebu{$\pm$0.20}/ & 0.26\rebu{$\pm$0.21}/  \\
& +0.31\rebu{$\pm$0.74} & 0.49\rebu{$\pm$0.79} &  +0.27\rebu{$\pm$0.62} & 0.47\rebu{$\pm$0.63} & +0.51\rebu{$\pm$0.62} & 0.62 \rebu{$\pm$0.58} & +0.52\rebu{$\pm$0.51} & 0.63\rebu{$\pm$0.51}  \\
\rule{0pt}{8pt}
\multirow{2}{*}{EWC}  & +0.10\rebu{$\pm$0.25/} & 0.19\rebu{$\pm$0.25}/ & +0.05\rebu{$\pm$0.13}/ & 0.17\rebu{$\pm$0.13}/ & +0.12\rebu{$\pm$0.17}/ & 0.22\rebu{$\pm$0.16}/ &  +0.10\rebu{$\pm$0.18}/& 0.21\rebu{$\pm$0.19}/  \\
        & +0.28\rebu{$\pm$0.67} &  0.46\rebu{$\pm$0.66} & +0.12\rebu{$\pm$0.37} & 0.37\rebu{$\pm$0.35} & +0.33\rebu{$\pm$0.44} & 0.51\rebu{$\pm$0.41}  & +0.27\rebu{$\pm$0.47} & 0.48\rebu{$\pm$0.45}  \\
\rule{0pt}{8pt}
\multirow{2}{*}{\rebu{Coreset}}  & \rebu{+0.03$\pm$0.09/ }& \textbf{\rebu{0.16$ \pm$0.12/ }}&\rebu{ +0.05$\pm$0.11/ }&\rebu{ 0.17$\pm$0.14/ }&\rebu{ +0.03$\pm$0.10/ }&\rebu{ 0.17$\pm$0.13}/ &\rebu{  \textbf{+0.04$\pm$0.09}/}&\rebu{ 0.19$\pm$0.15/}  \\
        &\rebu{ +0.09$\pm$0.27}&\textbf{\rebu{  0.36$\pm$0.34 }}&\rebu{ +0.12$\pm$0.29 }&\rebu{ 0.38$\pm$0.34 }&\rebu{ +0.08$\pm$0.25 }&\rebu{ \textbf{0.36$\pm$0.28}  }&\rebu{ +0.12$\pm$0.24}&\rebu{ 0.40$\pm$0.31}  \\
\rule{0pt}{8pt}
\multirow{2}{*}{SCL}  & \textbf{+0.02\rebu{$\pm$0.10/}} & 0.16\rebu{$\pm$0.14/} & \textbf{+0.01\rebu{$\pm$0.08/}} & \textbf{0.15\rebu{$\pm$0.12/}} & \textbf{+0.03\rebu{$\pm$0.08/}} & \textbf{0.17\rebu{$\pm$0.12/}} &  \textbf{+0.04\rebu{$\pm$0.09/}} &  \textbf{0.17\rebu{$\pm$0.13/}} \\
        & \textbf{+0.07\rebu{$\pm$0.29}} &  0.36\rebu{$\pm$0.40} & \textbf{+0.04\rebu{$\pm$0.20}} & \textbf{0.34\rebu{$\pm$0.27}} & \textbf{+0.07\rebu{$\pm$0.20}} & 0.37\rebu{$\pm$0.30}  & \textbf{+0.08\rebu{$\pm$0.21}} & \textbf{0.36\rebu{$\pm$0.28}}  \\
\hline
\rule{0pt}{7pt}
\multirow{2}{*}{\rebu{SCL-10}}  &\rebu{ +0.00$\pm$0.10/ }&\rebu{ 0.20$\pm$0.17/ }&\rebu{ +0.01$\pm$0.12/ }&\rebu{ 0.20$\pm$0.16/ }&\rebu{ +0.00$\pm$0.08/ }&\rebu{ 0.20$\pm$0.15/ }&\rebu{  +0.01$\pm$0.10/ }&\rebu{  0.20$\pm$0.14/} \\
        &\rebu{ +0.00$\pm$0.23 }&\rebu{ 0.45$\pm$0.40 }&\rebu{ +0.03$\pm$0.25 }&\rebu{ 0.44$\pm$0.37 }&\rebu{ +0.01$\pm$0.19 }&\rebu{ 0.45$\pm$0.33}&\rebu{ +0.01$\pm$0.26 }&\rebu{ 0.44$\pm$0.33} \\
\rule{0pt}{10pt}
\multirow{2}{*}{\rebu{SCL-20}}  &\rebu{ +0.04$\pm$0.12/ }&\rebu{ 0.22$\pm$0.19/ }&\rebu{ +0.02$\pm$0.10/ }&\rebu{ 0.20$\pm$0.16/ }&\rebu{ +0.04$\pm$0.11/ }&\rebu{ 0.21$\pm$0.18/ }&\rebu{  +0.03$\pm$0.12/ }&\rebu{  0.22$\pm$0.18/} \\
        &\rebu{ +0.10$\pm$0.31 }&\rebu{ 0.49$\pm$0.42 }&\rebu{ +0.05$\pm$0.25 }&\rebu{ 0.46$\pm$0.37 }&\rebu{ +0.08$\pm$0.26 }&\rebu{ 0.47$\pm$0.40}&\rebu{ +0.06$\pm$0.28 }&\rebu{ 0.50$\pm$0.41} \\
\hline 
\end{tabular}
\end{table*}}

\begin{figure}[t]
    \centering
    \adjustbox{trim={.0\width} {0.05\height} {0.05\width} {0.05\height},clip}{\includegraphics[width=0.44\textwidth]{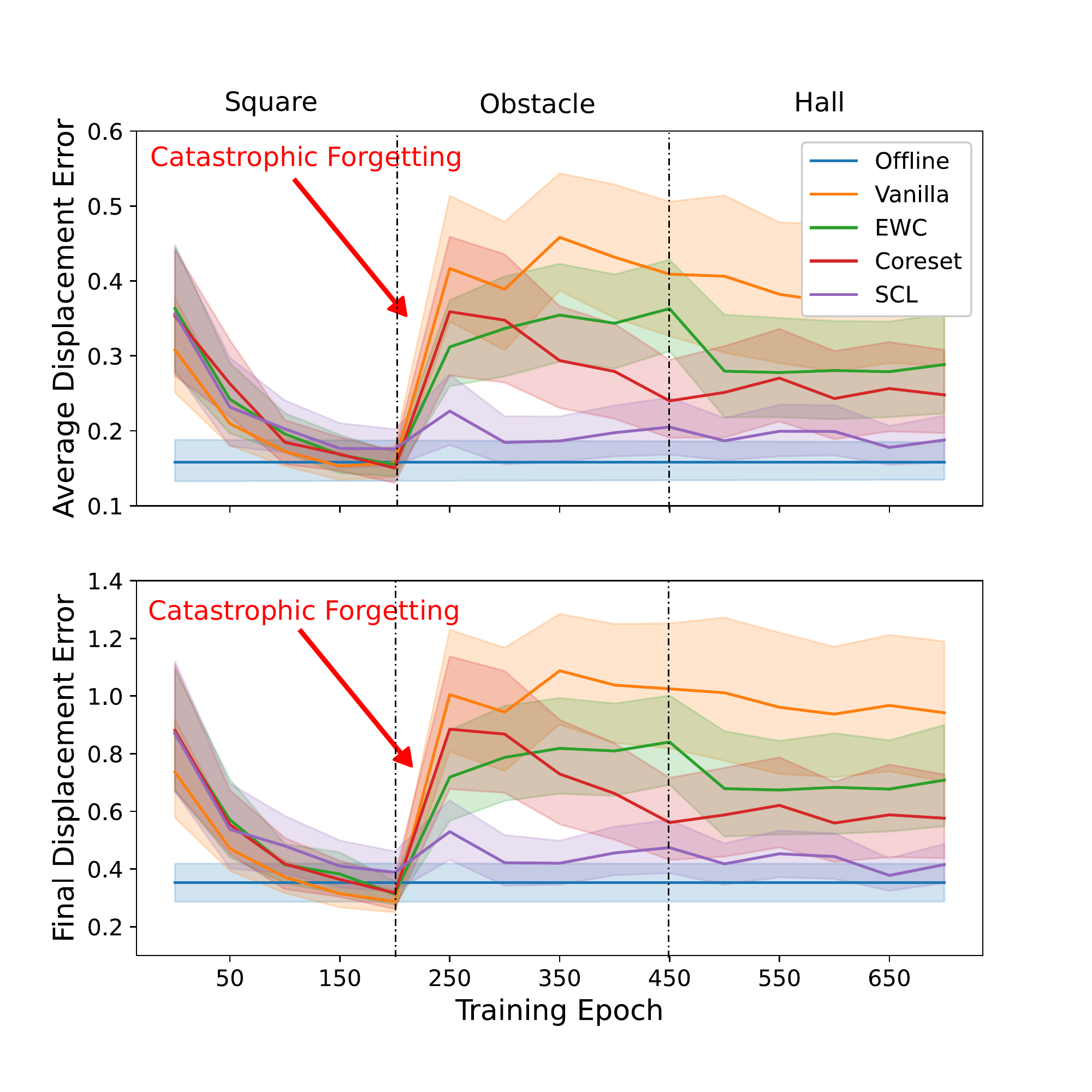}}
    \caption{\rebu{Prediction performance of models trained on square $\rightarrow$ obstacle $\rightarrow$ hall sequence evaluated on \textbf{only} the square scenario. 
    Shows ADE (top) and FDE (bottom) of all training methods on the validation set of the square task, while learning new tasks. 
    Note that the offline model is added for comparison purposes only.}}
    \label{fig:square}
\end{figure} 

\Cref{table:orders-sim} reports the \blue{mean and standard deviation (std) of} ADE/FDE evaluated at the end of the sequence on all three environments, under the columns denoted by \blue{\textit{seq.~end}}. The columns denoted by \blue{\textit{forgotten}} report the \blue{mean and std of} ADE/FDE increase of the prediction model on previous environments after training on new environments. It can be seen that SCL outperforms Vanilla. The significant increase in mean forgotten ADE/FDE for Vanilla indicates that naive online training over changing environments using standard gradient descent results in catastrophic forgetting. Our method is independent of sequence order, arriving at within $\pm 0.02$ of the same mean ADE/FDE for all orders.

To gain insight into where catastrophic forgetting occurs, we save the models trained for the sequence (square $\shortrightarrow$ obstacle $\shortrightarrow$ hall) after each training step and apply them to the \blue{validation} set of the square scenario only. \blue{\Cref{fig:square}} compares the performance of the different training methods on the square scenario \blue{validation} set at each training step. By evaluating a single environment over time, we can clearly visualize when and how much the models degraded in prediction performance in the respective environment. 
For ease of comparison, the offline trained model is also plotted as a constant line.
In the first section, all models are trained on the aggregated dataset of the square scenario and, as expected, \rebu{the error measures decrease for all online learning methods reaching better performance than offline trained prediction model due to overfitting}. However, when changing from the square environment to the obstacle environment, the ADE/FDE performance quickly and drastically degrades for the Vanilla baseline (red arrow). It can be seen that using EWC to selectively slow down learning on important parameters helps to significantly mitigate the magnitude of the loss in ADE/FDE. Nevertheless, after two subsequent tasks, the EWC baseline performed $\sim 30\%$ worse on FDE and $\sim 20\%$ worse on ADE.
\rebu{Rehearsing a set of past examples enables to retain more knowledge after two subsequent tasks than applying EWC.} Combining EWC and the coreset rehearsal as done in SCL helps to further mitigate forgetting. SCL was able to train in two subsequent scenarios while retaining knowledge of the initially experienced scenario.\\
\rebu{We have performed pair-wise Mann-Whitney U tests between our proposed method and each baseline to evaluate the statistical significance of the presented results. \cref{table:significance} shows the $p$-values comparing the performance results (i.e., ADE and FDE) on each scenario for the obstacle $\shortrightarrow$ hall $\shortrightarrow$ square sequence.}
\rebu{SCL significantly outperforms CV on all environments, the Vanilla baseline on all past environments, and EWC on one environment. 
Rehearsing alone achieves marginally worse results than SCL. Please note that the presented results consider a limited set of environments with limited complexity. We expect that as the number of scenarios and complexity increases, differences in performance between the baselines become significant.}

\renewcommand{\arraystretch}{0.9}{
\begin{table}[!t]
\setlength\tabcolsep{3.0pt}
\caption{\rebu{Statistical Significance Analysis using the Mann-Whitney U test. Comparison of SCL's performance (i.e., ADE and FDE) against all baselines on each environment for the obstacle $\shortrightarrow$ hall $\shortrightarrow$ square sequence. Significant results are displayed in bold considering a 5\% confidence-level.}}
\centerline{
\begin{tabular}{|p{0.115\linewidth}||>{\centering\arraybackslash}p{0.115\linewidth} | >{\centering\arraybackslash}p{0.115\linewidth}||>{\centering\arraybackslash}p{0.115\linewidth}|>{\centering\arraybackslash}p{0.115\linewidth}|| >{\centering\arraybackslash}p{0.115\linewidth}|>{\centering\arraybackslash}p{0.115\linewidth}|}
\hline
\rule{0pt}{7pt}
& \multicolumn{2}{c||}{obstacle} & \multicolumn{2}{c||}{hall} & \multicolumn{2}{c|}{square}\\
\hline
\rule{0pt}{7pt}
Method & ADE  & FDE  & ADE & FDE & ADE & FDE \\
\hline
\rule{0pt}{7pt}
CV & \textbf{p = 0.00} & \textbf{p = 0.00} & \textbf{p = 0.00} & \textbf{p = 0.00} & \textbf{p = 0.00} & \textbf{p = 0.00}\\
\rule{0pt}{7pt}
Vanilla & \textbf{p = 0.00} & \textbf{p = 0.00} & \textbf{p = 0.00} & \textbf{p = 0.00} & p = 0.97 & p = 0.53\\
\rule{0pt}{7pt}
EWC & p = 0.06 & \textbf{p = 0.02} & \textbf{p = 0.04} & \textbf{p = 0.01} & p = 0.65 & p = 0.71 \\
\rule{0pt}{7pt}
Coreset & p = 0.29 & p = 0.09 & p = 0.63 & p = 0.68 & p = 0.22 & p = 0.33\\
\hline
\end{tabular}
}
\label{table:significance}
\end{table}}

\subsection{\blue{Dense Scenarios}}
\blue{To evaluate how well our framework scales to complex scenarios with more pedestrians we employ the above simulation environments with increased numbers of pedestrians ($n= \{10,20\}$). The results are presented in \cref{table:orders-sim}.} \blue{It can be seen that SCL scales well to dense scenarios with more agents achieving similar performance for $10$ and $20$ pedestrians. The forgotten ADE/FDE even decreases for some sequences, indicating that observing more pedestrians can improve the preservation of past experiences.}

\subsection{Real-world Results}
We first evaluate our method in real-world experiments assuming perfect perception capabilities by using an external high-precision Optitrack tracking system. Secondly, we use the robot's on-board sensing capabilities combined with a detection and tracking pipeline.

\subsubsection{Perfect Perception}
\begin{figure}[t]
    \adjustbox{trim={.0\width} {0\height} {0.05\width} {0.15\height},clip}{\includegraphics[width=0.45\textwidth]{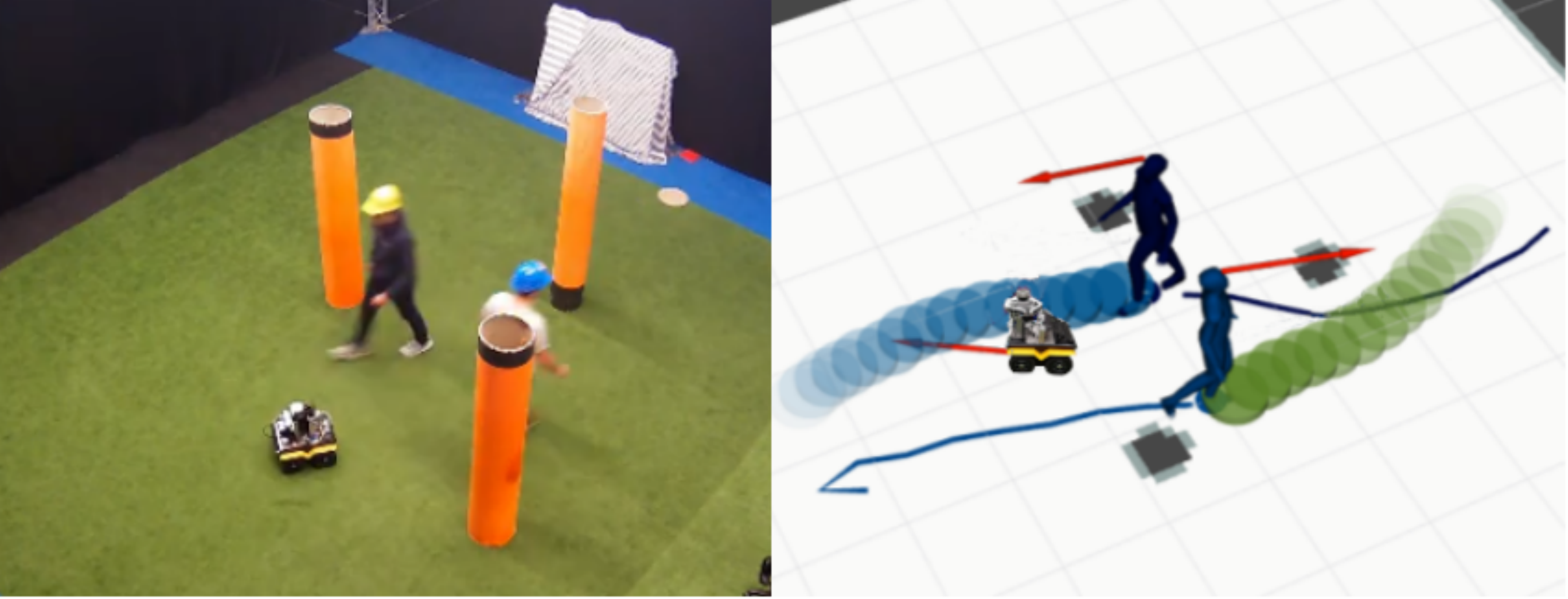}}
    \centering
    \caption{Real-world validation using an Optitrack system that streams the pedestrian states. The predicted pedestrian trajectories are depicted as green and blue disks.}
    \label{fig:qual_cyberzoo}
\end{figure}

To evaluate our framework using the Optitrack system, we set up three environments to replicate the ones considered in simulation (i.e., square, obstacle, cooperative). Each environment is observed for $T$ seconds.
\Cref{table:orders-cyberzoo} reports quantitative results on two different sequence orders similar to \cref{table:orders-sim}.
Our framework significantly outperformed the Vanilla baseline on both metrics indicating that we can not only learn a prediction model from real human motion but also that we need to consolidate the learned knowledge. SCL was able to improve prediction performance and learn certain concepts, such as avoiding crashing into walls, pillars, or fences. \Cref{fig:qual_cyberzoo} shows a qualitative example of the experiment, where our framework learns to avoid both static obstacles and pedestrians.

\subsubsection{On-board Perception}
\begin{figure}[t]
    \centering
    \adjustbox{trim={.0\width} {0.15\height} {0.00\width} {0.10\height},clip}{\includegraphics[width=0.40\textwidth]{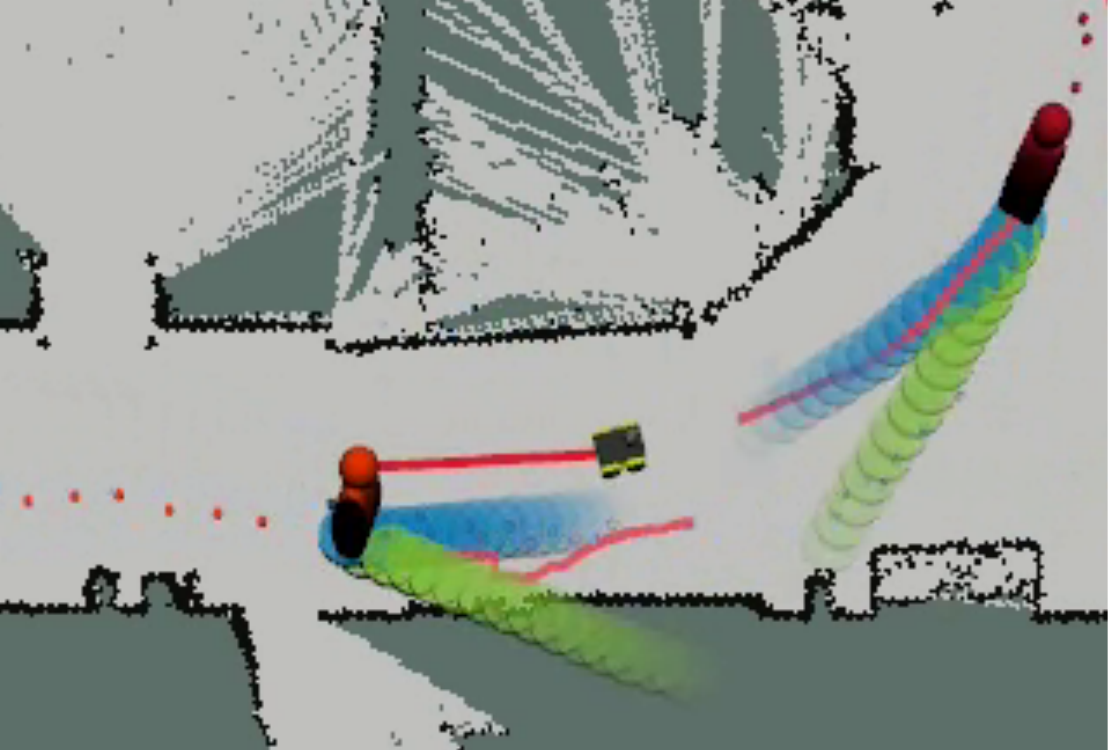}}
    \caption{\rebu{Map view of the real-world application of SCL on moving robot using on-board perception. The green and blue disks depict the predicted trajectories employing the pre-trained model and the SCL-trained model, respectively. The red dotted lines depict the pedestrians' past trajectories. The pedestrians' and robot's future trajectories are shown as solid red lines.}} 
    \label{fig:qual_3me}
\end{figure}

We now evaluate our framework in an uncontrolled hall environment using the robot's detection and tracking pipeline \blue{(i.e., LiDAR and cameras)}. \blue{In \cref{fig:qual_3me} we show qualitative results of the experiments with a moving robot.} 
\blue{The fact that the robot is constantly moving reduced the average collected trajectory length of the interacting pedestrians making the prediction problem harder. Thus, employing SCL in more dense environments is expected to further improve the resulting prediction performance.}
Nevertheless, the prediction model learned online when pedestrians are likely to take corners, by observing how real pedestrians walk in the environment. Note that the ETH and UCY datasets, on which our model was pre-trained, contain almost no interactions with static obstacles, yet our framework autonomously learns obstacle interactions. Furthermore, the occupancy map shown in \cref{fig:qual_3me} is generated by the robot itself using the depth information from its LiDAR. Thus, our framework can continuously learn in new and unseen environments autonomously. 

\begin{table}[!t]
\setlength\tabcolsep{3.0pt}
\caption{Quantitative results of Vanilla, EWC and SCL on real-world data collected using an optical tracking system. \blue{The table lists the mean$\pm$standard deviation (std) of ADE / FDE on all environments at the sequence end under \textit{seq. end}} and the \blue{mean$\pm$std} of the \textit{forgotten} ADE / FDE, which refers to the average increase in ADE / FDE on previous environments across the learning sequence. \blue{All error measures are presented in meters.}}
\centerline{
\begin{tabular}{|l || l | l || l | l |}
\hline
\rule{0pt}{7pt}
& \multicolumn{2}{c||}{square $\shortrightarrow$ obstacle $\shortrightarrow$ coop.} & \multicolumn{2}{c|}{obstacle $\shortrightarrow$ square $\shortrightarrow$ coop.}\\
\hline
 Method              & \thead{forgotten \\ \blue{(mean$\pm$std)}}  & \thead{ seq. end \\ \blue{(mean$\pm$std)}} & \thead{forgotten \\ \blue{(mean$\pm$std)}}  & \thead{seq. end\\ \blue{(mean$\pm$std)}} \\
\hline
\rule{0pt}{7pt}
\multirow{2}{*}{Vanilla} & +0.24\rebu{$\pm$0.28}/  & 0.46\rebu{$\pm$0.29}/ & +0.21\rebu{$\pm$0.26}/   & 0.45\rebu{$\pm$0.29}/  \\
& +0.58\rebu{$\pm$0.67} & 0.97\rebu{$\pm$0.66}  & +0.50\rebu{$\pm$0.64}& 0.94\rebu{$\pm$0.63}\\
\rule{0pt}{8pt}
\multirow{2}{*}{EWC}  & +0.19\rebu{$\pm$0.29}/  & 0.43\rebu{$\pm$0.27}/  & +0.12\rebu{$\pm$0.23}/ & 0.41\rebu{$\pm$0.25}/\\
& +0.42\rebu{$\pm$0.67}& 0.86\rebu{$\pm$0.61}& +0.31\rebu{$\pm$0.58} & 0.87\rebu{$\pm$0.56} \\
\rule{0pt}{8pt}
\multirow{2}{*}{SCL} & \textbf{+0.04\rebu{$\pm$0.21/}} & \textbf{0.36\rebu{$\pm$0.23/}} & \textbf{+0.05\rebu{$\pm$0.22/}} & \textbf{0.40\rebu{$\pm$0.28/}} \\
& \textbf{+0.13\rebu{$\pm$0.50}} & \textbf{0.73\rebu{$\pm$0.56}} & \textbf{+0.11\rebu{$\pm$0.50}} & \textbf{0.80\rebu{$\pm$0.60}}\\
\hline
\end{tabular}
}
\label{table:orders-cyberzoo}
\end{table}

%% file: sections/6_conclusions.tex
\section{CONCLUSIONS \& FUTURE WORK}
This paper introduces a Self-supervised Continual Learning framework (SCL) to improve pedestrian prediction models using online streams of data. We combined Elastic Weight Consolidation (EWC) and the rehearsal of a small constant sized set of examples to overcome catastrophic forgetting. 
We showed through experiments that SCL significantly outperforms vanilla gradient descent and performs similarly to offline trained models with full access to pedestrian data in all considered environments. Additionally, we showed in real-world experiments that our pedestrian prediction model can learn to generalize to new and unseen environments over time. Future work can investigate different methods to determine when the model should be updated, \blue{how different pedestrian behaviour types could be integrated into our framework} and the integration of our approach with a motion planner to improve the interaction-awareness 
between pedestrians and the robot.